\documentclass{article}

\usepackage{arxiv}

\usepackage[utf8]{inputenc}
\usepackage[T1]{fontenc}
\usepackage{hyperref}
\usepackage{url}
\usepackage{booktabs}
\usepackage{amsfonts}
\usepackage{nicefrac}
\usepackage{microtype}
\usepackage{relsize}
\usepackage{verbatim}
\usepackage{graphicx}
\usepackage{dcolumn}
\usepackage{bm}
\usepackage{float}
\usepackage{graphics}
\usepackage{color}
\usepackage{xcolor}
\usepackage{soul}
\usepackage{subfigure}
\usepackage{subcaption}
\usepackage{tikz}
\usetikzlibrary{arrows.meta,positioning,fit,calc}
\usepackage{booktabs}
\usepackage{amsmath,amssymb,amsfonts}

\usepackage{algorithm}
\usepackage{algpseudocode}
\usepackage{enumitem}
\setlist[itemize]{leftmargin=*}
\setlist[enumerate]{leftmargin=*}
\definecolor{rev}{rgb}{0,0,0}
\definecolor{rev2}{rgb}{0,0,0}
\usepackage{array}
\newcolumntype{P}[1]{>{\centering\arraybackslash}p{#1}}
\usepackage{multirow}
\usepackage{cancel}
\usepackage[capitalise]{cleveref}

\usepackage{cite}
\usepackage{tabularx}
\usepackage{threeparttable}
\usepackage{pifont}

\newcolumntype{Y}{>{\centering\arraybackslash}X}

\title{WLNO: Wavelet-Laplace Neural Operator for Solving Partial Differential Equations}

\author{
  Muhammad Abid \\
  Department of Mechanical and Aerospace Engineering,\\
  University of Tennessee, Knoxville\\
  Knoxville, TN 37996, USA.\\
  \texttt{mabid@vols.utk.edu}
  \And
  Arth Sojitra\\
  Department of Mechanical and Aerospace Engineering,\\
  University of Tennessee, Knoxville\\
  Knoxville, TN 37996, USA.\\
  \texttt{asojitra@vols.utk.edu}\\
  \And
  Omer San \\
  Department of Mechanical and Aerospace Engineering,\\
  University of Tennessee, Knoxville\\
  Knoxville, TN 37996, USA.\\
  \texttt{osan@utk.edu}
}

\begin{document}
\maketitle

\begin{abstract}
This work introduces the Wavelet-Laplace Neural Operator (WLNO), a novel neural operator that
fuses Haar wavelet multi-scale spatial decomposition with the Laplace-domain pole-residue
formulation of the Laplace Neural Operator (LNO). While LNO captures transient and steady-state
dynamics through learnable system poles and residues, it lacks an explicit mechanism for extracting
spatially localized multi-scale features inherent in complex PDE solutions. WLNO addresses this
by augmenting the LNO core with a parallel single-level Haar discrete wavelet transform (DWT)
branch that decomposes the lifted feature map into four frequency subbands: approximation (LL),
horizontal detail (LH), vertical detail (HL), and diagonal detail (HH) and applies independent
learned $1\times1$ convolutions to each subband before reconstruction via the inverse DWT. The
two branches are fused through a learnable sigmoid-gated weight $\alpha_\mathrm{wav}$,
initialized to give a small initial contribution to the wavelet branch, allowing the model to
adaptively balance Laplace-domain dynamics against spatial multi-scale features throughout
training. WLNO is evaluated against LNO on five benchmark PDE problems using identical
hyperparameters, training data, and evaluation protocols: the diffusion equation, the Burgers
equation, the reaction-diffusion system, Darcy flow, and the two-dimensional Navier-Stokes
equation. WLNO consistently outperforms LNO on all five problems with the most pronounced improvement on problems with strong spatial multi-scale
structure, such as the Burgers equation with sharp shock fronts and the Navier-Stokes equation
with coherent vortical structures, while remaining consistent across smoother and elliptic
problems. These results demonstrate that wavelet-based multi-scale spatial decomposition is a
principled and effective complement to Laplace-domain operator learning.
\end{abstract}


%=============================================================
\section{Introduction}
\label{sec:intro}
%=============================================================

The numerical solution of partial differential equations (PDEs) serves as the fundamental method for solving various problems that arise in computational science and engineering fields from fluid dynamics to structural mechanics and heat transfer and reaction chemistry~\cite{evans2010partial, leveque2002finite, quarteroni2008numerical}. The traditional numerical solvers, which include finite element methods and finite difference methods and finite volume methods, deliver precise results. However, their high computational demands make them costly to operate because any small modifications to boundary conditions or forcing functions or material parameters require complete resolution of the entire discretized system, which renders real-time operations and multiple query systems unfeasible according to~\cite{hesthaven2016certified, rozza2008reduced}. The computational bottleneck becomes most severe when applying parametric studies and uncertainty quantification and design optimization and digital twin applications, which require the solution of thousands of partial differential equations. The field of reduced-order modeling has advanced because this research established a foundation that includes proper orthogonal decomposition~\cite{benner2015survey} and non-intrusive regression approaches~\cite{hesthaven2018non} although traditional techniques face two main constraints. Recent developments have additionally explored solver-free and data-efficient training paradigms for neural operators~\cite{hasani2024generating,bischof2026hypino,sojitra2025method}.

The emergence of deep learning has developed a completely new research path that enables direct learning of solution operators through data analysis~\cite{lecun2015deep, goodfellow2016deep}. Neural networks function as surrogate models, which, after completing their offline training with numerical solver input-output data, can perform online solution evaluations within milliseconds without needing to be retrained~\cite{dissanayake1994neural, lagaris1998artificial}. This data-driven approach has demonstrated great success when applied to neural operator architectures, which learn to map between infinite-dimensional function spaces instead of mapping between finite-dimensional vector spaces~\cite{kovachki2023neural}. Unlike classical surrogate models, neural operators demonstrate their ability to maintain precise results across all spatial resolutions because their training process allows evaluation at any degree of spatial resolution without the need for subsequent training. The two studies~\cite{karniadakis2021physics, cuomo2022scientific} show complete assessments of scientific machine learning and physics-informed methods, while the study~\cite{peng2021multiscale} examines the complete problem of scale integration in computational science.

Two essential neural operator architectures have made their most significant impact on the field. The Deep Operator Network (DeepONet) ~\cite{lu2021learning}, represents the solution operator as a sum of products of two subnetworks: a branch network encoding the input function and a trunk network encoding the output coordinates. The Fourier Neural Operator (FNO)~\cite{li2020fourier} operates within the Green's function framework through its method of direct kernel parameterization that works in Fourier space to achieve fast computations through fast Fourier transform (FFT). FNO and its variants~\cite{li2020multipole, wen2022u, li2022fourier, tran2023factorized} have achieved state-of-the-art accuracy on a wide range of PDE benchmarks, and they successfully operate for weather prediction~\cite{pathak2022fourcastnet}, turbulence modeling~\cite{li2022fourier}, and seismic wave simulation~\cite{yang2021seismic}. Fourier-based operators achieve success yet face a fundamental limitation because the Fourier transform requires periodic signals while it remains unable to handle transient responses and initial conditions. The recently proposed Laplace Neural Operator (LNO) solves these problems through its implementation of a Fourier integral kernel that the Laplace domain operator uses to define its pole-residue method ~\cite{cao2023lno}. LNO learns both transient and steady-state responses simultaneously through trainable system poles $\mu_n \in \mathbb{C}$ and residues $\beta_n \in \mathbb{C}$, and has demonstrated that a single Laplace layer can outperform four Fourier modules across the different PDE benchmarks. 
The LNO method effectively captures time-based changes but fails to model spatial patterns because it uses global Fourier decomposition, which processes all spatial data yet lacks a dedicated system to trace different spatial scales of PDE solutions.

A complementary approach to handling multi-scale features in PDEs is the use of wavelet transforms~\cite{daubechies1992ten, mallat1999wavelet}, which provide representations that are simultaneously localized in both space and frequency. Unlike the Fourier transform, which represents signals as globally supported sinusoids, wavelets decompose signals into spatially localized basis functions at multiple scales, making them particularly effective for representing sharp gradients, shock fronts, and spatially heterogeneous features. The Wavelet Neural Operator (WNO)~\cite{tripura2023wavelet} has demonstrated that replacing Fourier convolutions with wavelet convolutions improves accuracy on problems with sharp features, at the cost of discarding the transient-response and initial-condition-handling capabilities of Laplace-based methods. This motivates the question at the core of the present work: can one design a neural operator that inherits both LNO's temporal dynamics representation and the wavelet's spatial multi-scale decomposition, and do these two capabilities provide genuinely complementary rather than redundant information?

In this work, we propose WLNO (Wavelet-Laplace Neural Operator), which provides an affirmative answer to this question. The key insight is that LNO's Laplace-domain computation captures the temporal dynamics of PDE solutions transient decay and steady-state oscillations controlled by the system poles $\mu_n$ while a wavelet branch independently captures the spatial multi-scale structure of those solutions by decomposing the lifted feature map into frequency-oriented subbands and learning independent transformations at each spatial frequency. These two operations target different dimensions of the PDE solution (time versus space, global versus local) and are therefore complementary rather than redundant. Their fusion via a learnable sigmoid-gated scalar weight $\alpha_\mathrm{wav}$ allows the model to adaptively allocate representational capacity between the two branches as training progresses, starting from a configuration where the PR2d branch dominates due to the small initialization and gradually activating the wavelet branch as the model learns to exploit spatial multi-scale features. 

Our main contributions are: (i) we propose WLNO, augmenting LNO's pole-residue Laplace layer
with a parallel single-level Haar wavelet branch fused via a learnable sigmoid-gated weight
$\alpha_\mathrm{wav}$; (ii) we provide a systematic empirical evaluation on five PDE benchmarks, the diffusion equation, the Burgers equation, the reaction-diffusion system, Darcy flow, and
the two-dimensional Navier-Stokes equation, using the same hyperparameters and data splits as
the original LNO paper, demonstrating consistent improvement over LNO with relative
$\mathcal{L}_2$ error reductions of $32.3\%$, $22.8\%$, $14.1\%$, $12.9\%$, and $20.7\%$,
respectively; and (iii) we analyze the learned fusion weights after training and show that the
wavelet branch becomes genuinely active, with $\alpha_\mathrm{wav}$ growing substantially from
its initialization of $0.12$, confirming that both branches are exploited by the network across
all benchmarks.

The remainder of this paper is organized as follows. Section~\ref{sec:related} reviews related work on neural operators, wavelet-based methods, and physics-informed approaches. Section~\ref{sec:method} presents the WLNO architecture in full detail, covering the LNO background, the Haar wavelet decomposition, the wavelet integral operator, the WLNO layer and fusion formulation, the full architecture, and the training algorithm. Section~\ref{sec:experiments} presents the experimental setup, results, and discussion for all five PDE benchmarks, including training curves, prediction visualizations, and quantitative comparisons, followed by a section-level summary. Section~\ref{sec:conclusion} provides the overall summary, and Section~\ref{sec:future} outlines directions for future work.

%=============================================================
\section{Related Work}
\label{sec:related}
%=============================================================
Neural operators serve as an expansion of neural networks, which transform finite-dimensional mappings into functional space operator systems because of their ability to maintain evaluation accuracy across all spatial resolutions after their initial training~\cite{kovachki2023neural, bhattacharya2021model}. The universal approximation theorem for operators established by Chen and Chen~\cite{chen1995universal} serves as the primary theoretical foundation because it proves that any continuous nonlinear operator can be approximated with unlimited precision through DeepONet~\cite{lu2021learning} when specific conditions are met. Extensions of DeepONet include physics-informed variants~\cite{wang2022improved}, multifidelity formulations~\cite{de2022multifidelity}, spectrally-embedded trunks~\cite{sojitra2025fedonet,abid2025spectral} and uncertainty quantification approaches~\cite{lin2021operator}. FNO~\cite{li2020fourier} delivers its kernel integral operator through its Fourier space
convolution, which achieves $O(N \log N)$ processing using FFT. The initial FNO has produced
multiple derivatives that include GeoFNO~\cite{li2023geometryfno}, Tucker-FNO~\cite{tran2023factorized},
UFNO~\cite{wen2022u}, spherical FNO~\cite{bonev2023sfno}, and a spectrally-informed
multi-resolution extension for turbulent flow super-resolution~\cite{abid2026simr}. Attention-based architectures~\cite{vaswani2017attention} have been integrated into operator learning frameworks through their development into functional systems, whereas U-shaped encoder-decoder designs~\cite{ronneberger2015u} and residual connections~\cite{he2016deep} now function as training stabilization tools for deep operator stacks~\cite{kissas2022learning}. The Convolutional Neural Operator~\cite{raonic2024convolutional} utilizes convolutional layers that directly address aliasing artifacts, while Poseidon~\cite{herde2024poseidon} employs attention mechanisms to achieve scalable operator learning.

The Laplace Neural Operator (LNO)~\cite{cao2023lno} introduced a Laplace-domain pole-residue operator that substituted the original Fourier kernel, allowing the system to learn both transient and steady-state responses through its trainable complex-valued poles $\mu_n$ and residues, $\beta_n$ which were developed by the system. The Laplace Neural Operator (LNO)~\cite{cao2023lno} introduced a Laplace-domain pole-residue operator that substituted the original Fourier kernel, allowing the system to learn both transient and steady-state responses through its trainable complex-valued poles $\mu_n$ and residues, $\beta_n$ which were developed by the system. The central finding of LNO demonstrates that the network can identify essential dynamic modes of the PDE solution through data-driven learning of the previously established physical parameters known as poles $\mu_n$. The results in the development of a single-layer operator that demonstrates superior performance across all tests compared to four Fourier modules that were used in the original study~\cite{cao2023lno}. The Galerkin Transformer~\cite{cao2021choose} formulation from the spectral perspective replaces the quadratic attention with Fourier-space inner products, thereby increasing computational efficiency. Scientific computing has used frequency-domain representations since ancient times because these methods include both classical spectral techniques and current deep learning methods used in image processing and signal reconstruction~\cite{hendrycks2016gaussian}.

Wavelets deliver time-frequency analysis through their ability to maintain spatial and frequency localization~\cite{daubechies1992ten, mallat1999wavelet}, which enables them to accurately display signals that contain sudden changes and exhibit smooth areas and show different scale hierarchical patterns~\cite{chui1992introduction}. The scattering transform~\cite{mallat2012group} shows how deep convolutional networks process signals through a series of wavelet convolutions followed by modulus nonlinearities, which create a mathematical link between wavelet analysis and deep learning that existed before neural operators became common. The deep learning literature uses wavelets in CNNs for texture analysis according to~\cite{bruna2013invariant} and for multi-scale feature extraction according to~\cite{liu2019multi} and for image compression purposes. The proposed wavelet pooling system allows maintaining spatial details throughout different scales while avoiding aliasing problems that exist in traditional max-pooling methods~\cite{williams2018wavelet}. The Wavelet Neural Operator (WNO)~\cite{tripura2023wavelet} developed operator learning through their method, which used wavelet convolution instead of FNO's Fourier convolution and implemented learned lifting-based filters for their system. The system demonstrated better accuracy when solving problems that contained sharp solution features such as Burgers shocks and discontinuous initial conditions. The research uses multiwavelets, which deliver exact polynomial approximation while they possess multiple vanishing moments, as the foundation to develop integral operator learning. The developed approach shows better performance on smooth problems that exhibit high-order regularity~\cite{gupta2021multiwavelets}. Our method establishes a complete distinction from WNO and multiwavelet operators because we maintain the original integral kernel while introducing a wavelet component that operates together with the unaltered PR2d Laplace core. By trading one capability for another, it appears that it would have been a methodology for securing or further strengthening each one of LNO's demonstrated capabilities.

Physics-informed neural networks (PINNs)~\cite{raissi2019physics} use PDE residuals as their primary training loss component, which allows them to solve both forward and inverse problems when working with restricted data. The extensions of the system include adaptive sampling strategies~\cite{lu2021deepxde} and domain decomposition~\cite{jagtap2020extended} and hp-variational formulations~\cite{kharazmi2021hp} and applications to nonlinear fracture mechanics~\cite{goswami2022physics}. Neural operators provide an extension to PINNs because they acquire knowledge of solution mapping instead of directly obtaining the final solution~\cite{li2021physics}: a trained neural operator can handle various forcing functions and boundary conditions and initial conditions without needing a new training session, whereas a PINN requires complete resolution for every distinct problem case. Physics-informed neural operators (PINO)~\cite{li2021physics} combine both paradigms through their use of PDE residuals, which function as extra loss components during operator training to enhance their ability to generalize when operating with limited training data. The ability to handle chaotic time-dependent systems has been studied in the context of recurrent networks~\cite{vlachas2019data}, which created a need for operator-based methods that can handle multiple input situations without suffering from sequential prediction errors.

The linear subspace requirement together with the need for specific training snapshot distribution results in system limitations for both proper orthogonal decomposition~\cite{benner2015survey} and non-intrusive regression surrogates~\cite{hesthaven2018non} which represent classical reduced-order modeling techniques. The modern neural operators solve both existing restrictions because each operator uses its basic transformation to handle its fundamental drawbacks. The FNO model operates under the assumption of periodic behavior and maintains a constant structural state. The LNO system provides complete temporal analysis but fails to break down spatial data into multiple size categories. The WNO system identifies spatial features at different size dimensions, yet it does not function to track temporary changes. WLNO exists at this intersection point because it integrates Laplace-domain temporal dynamics with wavelet spatial multi-scale decomposition through a learnable adaptive weight to create a unified compact system that enables complete end-to-end training.

%=============================================================
\section{Wavelet-Laplace Neural Operator: Background, Architecture, and Training}
\label{sec:method}
%=============================================================

\subsection{Operator Learning and the LNO Framework}

Let $\Omega \subset \mathbb{R}^D$ be a bounded open domain and let $\mathcal{X}$, $\mathcal{Y}$ be separable Banach spaces on $\Omega$. The goal of operator learning is to approximate the nonlinear solution map $\mathcal{G}: \mathcal{X} \to \mathcal{Y}$ from labeled pairs $\{(f_j, u_j)\}_{j=1}^N$, where $f_j \in \mathcal{X}$ is a forcing function or initial condition and $u_j \in \mathcal{Y}$ is the corresponding PDE solution. The parametric approximation $\mathcal{G}_\theta: \mathcal{X} \to \mathcal{Y}$ is learned by minimizing the average relative error over the training set, and is required to generalize to unseen inputs drawn from a shifted distribution (different amplitude or decay rate of the forcing function).

LNO~\cite{cao2023lno} follows the standard neural operator pipeline with a crucial modification to the integral kernel. A lifting layer $\mathcal{P}$ maps the input $f(t) \in \mathbb{R}^{d_\mathrm{in}}$ to a higher-dimensional latent representation $v(t) \in \mathbb{R}^{d_z}$. A single Laplace layer then applies:
\begin{equation}
    u(t) = \sigma\!\left(\mathcal{K}_\phi(v)(t) + W v(t)\right),
    \label{eq:lno_layer}
\end{equation}
where $W$ is a local pointwise linear transformation (a $1\times1$ convolution serving as a residual bypass), $\mathcal{K}_\phi$ is the Laplace integral operator, and $\sigma$ is a nonlinear activation. A projection layer $\mathcal{Q}$ maps $u(t)$ back to the target dimension. The $\sin$ activation in the projection head~\cite{cao2021choose} and the GELU activation~\cite{hendrycks2016gaussian} used in the nonlinear correction branch are adopted from standard neural operator practice. The theoretical foundation of the pole-residue method draws from classical structural dynamics~\cite{kreyszig2010advanced, hu2016pole, cao2023laplace}.

The integral operator $\mathcal{K}_\phi$ is parameterized in the Laplace domain via the pole-residue method. The transfer function $K_\phi(s)$ is expressed as a partial-fraction sum:
\begin{equation}
    K_\phi(s) = \sum_{n=1}^{N} \frac{\beta_n}{s - \mu_n},
    \label{eq:transfer}
\end{equation}
where $\mu_n \in \mathbb{C}$ are the learnable system poles and $\beta_n \in \mathbb{C}$ are the learnable system residues. Both $\{\mu_n\}_{n=1}^N$ and $\{\beta_n\}_{n=1}^N$ are trainable complex parameters initialized from a scaled random distribution and updated end-to-end via backpropagation through the complex exponential computations. The real part of $\mu_n$ controls the temporal behavior of the associated mode: $\mathrm{Re}(\mu_n) < 0$ corresponds to a decaying transient, $\mathrm{Re}(\mu_n) > 0$ to an unstable growing mode, and $\mathrm{Re}(\mu_n) = 0$ to a purely oscillatory steady-state contribution. The imaginary part $\mathrm{Im}(\mu_n)$ controls the oscillation frequency of that mode.

The Laplace transform of the input $V(s) = \mathcal{L}\{v(t)\}$ is decomposed via its Fourier series as $V(s) = \sum_\ell \alpha_\ell / (s - i\omega_\ell)$, where $\alpha_\ell = \mathrm{FFT2}(v)|_\ell$ are the complex Fourier coefficients and $\omega_\ell = 2\pi\ell/T$ are the discrete angular frequencies. The product $U(s) = K_\phi(s)\,V(s)$ is then expanded via partial fractions into two physically meaningful components:
\begin{equation}
    U(s) = \underbrace{\sum_{n=1}^{N} \frac{\gamma_n}{s - \mu_n}}_{\text{transient (system poles)}} + \underbrace{\sum_{\ell} \frac{\lambda_\ell}{s - i\omega_\ell}}_{\text{steady-state (excitation poles)}},
    \label{eq:pole_residue}
\end{equation}
where the two sets of residues are computed analytically from the learned parameters:
\begin{equation}
    \gamma_n = \beta_n\, V(\mu_n) = \beta_n \sum_\ell \frac{\alpha_\ell}{\mu_n - i\omega_\ell}, \qquad
    \lambda_\ell = \alpha_\ell\, K_\phi(i\omega_\ell) = \alpha_\ell \sum_{n=1}^N \frac{\beta_n}{i\omega_\ell - \mu_n}.
    \label{eq:residues}
\end{equation}
The transient residues $\gamma_n$ couple the system poles to the input spectrum $\alpha_\ell$, encoding how strongly the $n$-th decaying/growing mode is excited by the forcing function at all frequencies. The steady-state residues $\lambda_\ell$ encode how the transfer function $K_\phi$ shapes the response at each excitation frequency $i\omega_\ell$. Taking the inverse Laplace transform of Eq.~\eqref{eq:pole_residue} gives the time-domain output:
\begin{equation}
    u(t) = \underbrace{\sum_{n=1}^{N} \gamma_n\, e^{\mu_n t}}_{u_\mathrm{tr}:\;\text{transient}} + \underbrace{\sum_\ell \lambda_\ell\, e^{i\omega_\ell t}}_{u_\mathrm{st}:\;\text{steady-state}}.
    \label{eq:lno_output}
\end{equation}
The steady-state term $u_\mathrm{st}$ is computed efficiently as an IFFT of the $\lambda_\ell$ spectrum, while the transient term $u_\mathrm{tr}$ is an explicit exponential sum evaluated at the output grid points. The complete PR2d output is $u_\mathrm{tr} + u_\mathrm{st}$. This formulation provides two decisive advantages over FNO: it represents exponentially decaying or growing transient responses through $\mathrm{Re}(\mu_n)$, and it captures the coupling between initial conditions and the forcing through $\gamma_n = \beta_n V(\mu_n)$, which mixes the system poles with the full input spectrum. Both of these capabilities are fundamentally inaccessible to FNO, which can only represent the steady-state periodic component $u_\mathrm{st}$.

\subsection{Motivation for Wavelet Augmentation}

Despite LNO's advantages over FNO in the temporal domain, a significant limitation remains on the spatial side. The PR2d operator acts on the lifted feature map $\mathbf{v} \in \mathbb{R}^{B \times d_z \times H \times W}$ through a global Fourier decomposition (the FFT2 in Eq.~\eqref{eq:residues}) and an exponential sum that couples all spatial locations via the system poles. The global spatial processing method works effectively with diffusion equations, which maintain smooth solutions that extend uniformly across space, but fails to handle partial differential equations that display spatial patterns throughout different dimensional ranges. The Burgers equation solution establishes shock front development through its creation of narrow, high-amplitude gradient peaks that exist at particular points in space. The Fourier representation requires multiple modes to capture these features, which exhibit non-local characteristics. The reaction-diffusion equation creates high-concentration areas that develop in specific locations because the quadratic reaction term cannot be accurately represented through Fourier modes which extend throughout all space.

Wavelets provide the natural remedy: by decomposing the feature map into spatially localized subbands at different frequency orientations, a wavelet branch can represent sharp edges, localized gradients, and multi-scale spatial structure much more compactly than a Fourier basis. Crucially, this spatial multi-scale decomposition is orthogonal to the temporal dynamics captured by the Laplace layer: the wavelet branch processes the spatial structure of the features at a single time step, while the PR2d branch handles how those features evolve over time through the system poles $\mu_n$. The two operations therefore target different aspects of the PDE solution and can be combined additively without redundancy.

\subsection{Haar Wavelet Decomposition}
\label{subsec:haar}

We employ the single-level 2D Haar DWT, implemented as fixed-weight strided convolutions without any external wavelet library~\cite{daubechies1992ten, mallat1999wavelet}. Define the low-pass analysis filter $h = [1,\,1]^\top/\sqrt{2}$ and the high-pass analysis filter $g = [1,\,-1]^\top/\sqrt{2}$. The four 2D separable analysis filters are constructed from outer products:
\begin{equation}
    \mathbf{h}_{kl} = \psi_k \otimes \psi_l \;\in\; \mathbb{R}^{2\times2},
    \qquad (k,l) \in \{(L,L),(L,H),(H,L),(H,H)\},
    \qquad \psi_L = h,\;\; \psi_H = g.
    \label{eq:filters}
\end{equation}
These four filters are registered as fixed (non-trainable) buffers in the network. For a feature map $\mathbf{x} \in \mathbb{R}^{B \times C \times H \times W}$, each channel is processed independently via strided convolution with stride 2:
\begin{equation}
    \mathbf{S}_{kl} = \mathbf{h}_{kl} *_{\mathrm{s}=2}\, \mathbf{x} \;\in\; \mathbb{R}^{B \times C \times H/2 \times W/2}, \qquad (k,l) \in \{L,H\}^2.
    \label{eq:dwt}
\end{equation}
The four subbands carry complementary spatial frequency content: $\mathbf{S}_{LL}$ is the low-frequency approximation capturing smooth global background; $\mathbf{S}_{LH}$ captures horizontal high-frequency edges (rapid variation in the vertical direction); $\mathbf{S}_{HL}$ captures vertical high-frequency edges (rapid variation in the horizontal direction); and $\mathbf{S}_{HH}$ captures diagonal high-frequency detail. The IDWT uses the same Haar filters in transposed convolutions to reconstruct the full-resolution map:
\begin{equation}
    \mathrm{IDWT}(\mathbf{S}_{LL}, \mathbf{S}_{LH}, \mathbf{S}_{HL}, \mathbf{S}_{HH})
    = \sum_{(k,l)\,\in\,\{L,H\}^2} \mathbf{h}_{kl}\,\bar{*}_{\mathrm{s}=2}\, \mathbf{S}_{kl}
    \;\in\; \mathbb{R}^{B \times C \times H \times W},
    \label{eq:idwt}
\end{equation}
where $\bar{*}_{\mathrm{s}=2}$ denotes the transposed convolution with stride 2. The Haar filters are orthogonal ($\mathbf{h}_{kl} = \mathbf{h}_{kl}^*$), so the analysis and synthesis filters are identical and perfect reconstruction is guaranteed. If the input has odd spatial dimensions, zero-padding is applied before the DWT and the extra row/column is cropped after IDWT to ensure exact size preservation.

\subsection{Wavelet Integral Operator}
\label{subsec:wavelet_op}

Given the four subbands $\{\mathbf{S}_{kl}\}$ of the normalized feature map, the wavelet integral operator applies an independent learned pointwise channel-mixing matrix $\mathbf{W}_{kl} \in \mathbb{R}^{C_\mathrm{out} \times C_\mathrm{in}}$ to each subband separately, then reconstructs the full-resolution output:
\begin{equation}
    \mathcal{W}(\mathbf{x}) = \mathrm{IDWT}\!\left(
    \mathbf{W}_{LL}\,\mathbf{S}_{LL},\;\;
    \mathbf{W}_{LH}\,\mathbf{S}_{LH},\;\;
    \mathbf{W}_{HL}\,\mathbf{S}_{HL},\;\;
    \mathbf{W}_{HH}\,\mathbf{S}_{HH}
    \right).
    \label{eq:wavelet_op}
\end{equation}
Each $\mathbf{W}_{kl}$ is a $1\times1$ convolution applied identically at every spatial location of the half-resolution subband. The four operators are independent, so the network can learn entirely different channel-mixing behaviors for each spatial frequency orientation: $\mathbf{W}_{LL}$ may learn to suppress noise in the smooth background while $\mathbf{W}_{HH}$ may learn to amplify diagonal edge features relevant to a specific PDE. This design has two important properties. First, it is lightweight: the four $1\times1$ convolutions on half-resolution subbands add only $4C^2$ parameters for $C_\mathrm{in} = C_\mathrm{out} = C$, compared to $C^2 k^2$ for a spatial convolution with kernel size $k$. Second, it is strictly non-mixing across orientations: the LL, LH, HL, and HH content is processed in separate streams before reconstruction, so the network cannot inadvertently confuse smooth background features with sharp edge content.

\subsection{WLNO Layer, Fusion, and Full Architecture}
\label{subsec:wlno_layer}

The WLNO layer receives the lifted feature map $\mathbf{v} \in \mathbb{R}^{B \times d_z \times H \times W}$ and routes it through two parallel branches, each normalized independently by instance normalization~\cite{ulyanov2016instance} to stabilize training:
\begin{align}
    \mathbf{p} &= \mathcal{K}_\phi\!\left(\mathrm{IN}(\mathbf{v})\right)
    \;\in\; \mathbb{R}^{B \times d_z \times H \times W},
    \label{eq:pr_branch}\\[2pt]
    \mathbf{w} &= \mathcal{W}\!\left(\mathrm{IN}_\mathrm{wav}(\mathbf{v})\right)
    \;\in\; \mathbb{R}^{B \times d_z \times H \times W},
    \label{eq:wav_branch}
\end{align}
where $\mathcal{K}_\phi$ is the PR2d Laplace operator (Eqs.~\eqref{eq:lno_output}--\eqref{eq:residues}) and $\mathcal{W}$ is the wavelet integral operator (Eq.~\eqref{eq:wavelet_op}). The two branch outputs are fused through a learnable scalar weight, and the fused result is added to a local pointwise linear bypass:
\begin{equation}
    \mathbf{u} = \mathrm{IN}\!\left(\mathbf{p} + \alpha_\mathrm{wav}\,\mathbf{w}\right) + \mathbf{W}_\mathrm{loc}\,\mathbf{v},
    \label{eq:fusion}
\end{equation}
where $\mathbf{W}_\mathrm{loc} \in \mathbb{R}^{d_z \times d_z}$ is a $1\times1$ Conv2d local bypass (the residual connection analogous to the $W$ term in LNO from Eq.~\eqref{eq:lno_layer}), and the outer $\mathrm{IN}$ is an additional InstanceNorm2d applied after fusion. The learnable fusion weight is parameterized via sigmoid gating:
\begin{equation}
    \alpha_\mathrm{wav} = \sigma(\ell_\mathrm{wav}), \qquad
    \sigma(z) = \frac{1}{1+e^{-z}}, \qquad
    \ell_\mathrm{wav} \in \mathbb{R}, \qquad
    \ell_\mathrm{wav}\big|_{t=0} = -2.0 \;\Rightarrow\; \alpha_\mathrm{wav}(0) \approx 0.12.
    \label{eq:alpha}
\end{equation}
The initialization $\ell_\mathrm{wav} = -2.0$ ensures that $\alpha_\mathrm{wav}$ starts near $0.12$, so the PR2d branch dominates early training and the network first develops a good Laplace-domain temporal representation before the wavelet branch gradually activates. This prevents the instability that would arise from a naive equal-weight initialization ($\ell_\mathrm{wav} = 0$, $\alpha_\mathrm{wav} = 0.5$) when the wavelet branch parameters $\{\mathbf{W}_{kl}\}$ have not yet been trained to produce meaningful features.

The complete WLNO architecture is illustrated in Fig.~\ref{fig:architecture}. The input function $f(\mathbf{x},t) \in \mathbb{R}$ is augmented with spatial grid coordinates $(\xi_1, \xi_2) \in [0,1]^2$ encoding location in the computational domain, and the concatenated vector is lifted to the latent channel space by a fully connected layer:
\begin{equation}
    \mathbf{v} = \mathcal{P}\!\left([f;\, \xi_1;\, \xi_2]\right)
    = \mathbf{W}_\mathcal{P}\,[f;\, \xi_1;\, \xi_2] + \mathbf{b}_\mathcal{P},
    \qquad \mathbf{W}_\mathcal{P} \in \mathbb{R}^{d_z \times 3},\;\; \mathbf{b}_\mathcal{P} \in \mathbb{R}^{d_z}.
    \label{eq:lifting}
\end{equation}
The lifted map $\mathbf{v}$ is permuted to channel-first format $(B, d_z, H, W)$, processed by the WLNO layer (Eq.~\eqref{eq:fusion}) to produce $\mathbf{u}$, permuted back to spatial-last format $(B, H, W, d_z)$, and projected to the scalar output field via a two-layer nonlinear projection head:
\begin{equation}
    \hat{y} = \mathbf{W}_2^\mathcal{Q}\,\sin\!\left(\mathbf{W}_1^\mathcal{Q}\,\mathbf{u}\right),
    \qquad \mathbf{W}_1^\mathcal{Q} \in \mathbb{R}^{128 \times d_z},\;\; \mathbf{W}_2^\mathcal{Q} \in \mathbb{R}^{1 \times 128}.
    \label{eq:projection}
\end{equation}
The $\sin$ nonlinearity in the projection head is adopted following~\cite{cao2021choose, cao2023lno} and provides smooth, periodic-compatible activations suited to PDE solutions. The grid coordinate augmentation (Eq.~\eqref{eq:lifting}) supplies the network with explicit spatial location information, which is essential for the PR2d branch to compute the correct exponential sum $e^{\mu_n t}$ at the right spatiotemporal coordinates and for the wavelet branch to produce spatially coherent multi-scale features.

\subsection{Architecture Diagram and Training Algorithm}
\begin{figure}[H]
    \centering
    \includegraphics[width=0.8\textwidth]{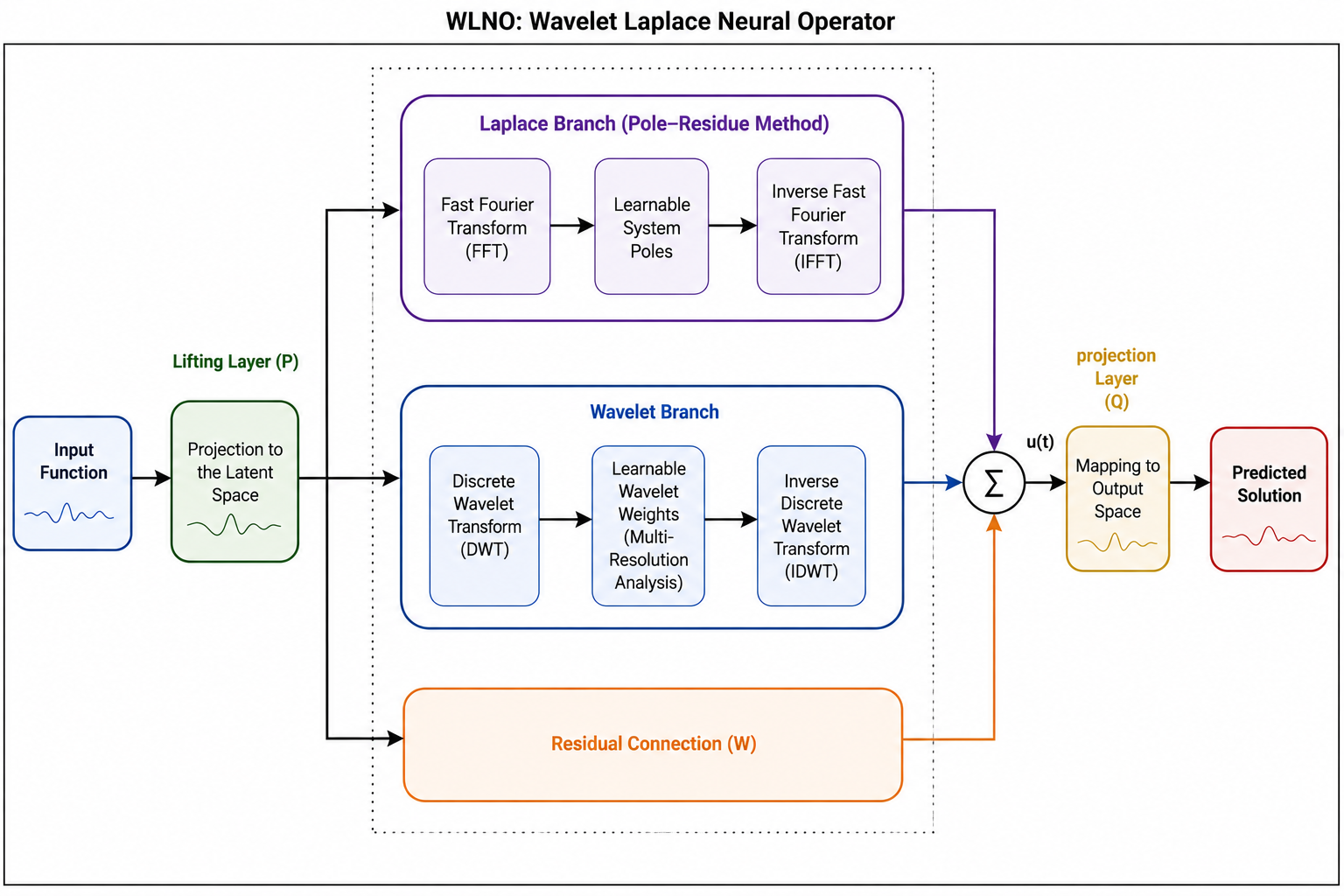}
    \caption{Architecture of WLNO. \textbf{(a)} Full pipeline: the input function $f(\mathbf{x},t)$ is concatenated with spatial grid coordinates $(\xi_1,\xi_2)$ and lifted to latent space $\mathbf{v} \in \mathbb{R}^{d_z}$ by $\mathcal{P}$. The WLNO layer processes $\mathbf{v}$ through two parallel branches, PR2d Laplace and Haar DWT wavelet, whose outputs are fused via the learnable weight $\alpha_\mathrm{wav}$ and added to the local bypass $\mathbf{W}_\mathrm{loc}$. The result $\mathbf{u}$ is projected to the output field $y(\mathbf{x},t)$ by $\mathcal{Q}$. \textbf{(b)} WLNO layer detail: instance-normalized inputs enter the PR2d branch (computing transient exponential sums and steady-state IFFT responses) and the WaveletLayer2d branch (Haar DWT $\to$ independent $1\times1$ convolutions per subband $\to$ IDWT) in parallel. Outputs are fused with learned weight $\alpha_\mathrm{wav}$, normalized, and added to the $1\times1$ Conv2d bypass.}
    \label{fig:architecture}
\end{figure}

The complete WLNO architecture is shown in Figure~\ref{fig:architecture}, which illustrates
the full pipeline and the internal design of the WLNO layer. The architecture takes an input
function, lifts it to a higher-dimensional latent space, processes it through the dual-branch
WLNO layer where the Laplace and wavelet branches operate in parallel, and projects the result
back to the output field. The two branch outputs are adaptively combined through the learnable
fusion weight $\alpha_\mathrm{wav}$ together with a local residual bypass before the final
projection.

The complete training procedure is described in Algorithm~\ref{alg:wlno}. The model is
initialized such that the Laplace branch dominates early training, allowing it to first
establish a reliable temporal representation before the wavelet branch gradually activates and
contributes its spatial multi-scale decomposition. The Haar wavelet filters are fixed throughout
training, while all other parameters including the fusion weight are jointly optimized
end-to-end, enabling the model to adaptively balance the contributions of the two branches as
training progresses.

\begin{algorithm}[H]
\caption{Training procedure for WLNO}
\label{alg:wlno}
\begin{algorithmic}[1]
\Require Training set $\mathcal{D} = \{(f^{(i)},\, y^{(i)})\}_{i=1}^N$,
         spatial grid $\{(\xi_1^q, \xi_2^q)\}_{q=1}^Q$,
         latent width $d_z$,
         learning rate $\eta$, weight decay $\lambda$,
         step size $S$, decay factor $\gamma$, total epochs $E$
\Ensure  Trained parameters $\theta = \{\mathbf{W}_\mathcal{P},\, \mu_n,\, \beta_n,\, \{\mathbf{W}_{kl}\},\, \ell_\mathrm{wav},\, \mathbf{W}_\mathrm{loc},\, \mathbf{W}_1^\mathcal{Q},\, \mathbf{W}_2^\mathcal{Q}\}$
\State Initialize $\mu_n, \beta_n$ from scaled complex random; all $\mathbf{W}$ from Kaiming uniform
\State Initialize $\ell_\mathrm{wav} \leftarrow -2.0$ \quad $\triangleright$ $\alpha_\mathrm{wav}(0) \approx 0.12$: PR2d branch dominates early
\State Build Haar filters $\{\mathbf{h}_{kl}\}$ from Eq.~\eqref{eq:filters} as fixed non-trainable buffers
\State Set up Adam optimizer with $\eta$, $\lambda$; StepLR scheduler decays by $\gamma$ every $S$ epochs
\For{$e = 1, \dots, E$}
    \For{each mini-batch $\mathcal{B} \subset \mathcal{D}$}
        \State Form $\tilde{f}^{(i)} = [f^{(i)};\, \xi_1;\, \xi_2] \in \mathbb{R}^{H \times W \times 3}$ for all $i \in \mathcal{B}$
        \State \textbf{Lifting:}\; $\mathbf{v}^{(i)} = \mathbf{W}_\mathcal{P}\,\tilde{f}^{(i)} + \mathbf{b}_\mathcal{P}$, permute to $(B, d_z, H, W)$ \hfill $\triangleright$ Eq.~\eqref{eq:lifting}
        \State \textbf{PR2d branch:}\; $\hat{\mathbf{v}} = \mathrm{IN}(\mathbf{v})$;\; $\alpha = \mathrm{FFT2}(\hat{\mathbf{v}})$;\; compute $\gamma_n$, $\lambda_\ell$ via Eq.~\eqref{eq:residues}
        \Statex \hspace{3.5em} $u_\mathrm{tr} = \mathrm{Re}\!\left(\sum_n \gamma_n e^{\mu_n t}\right)$;\; $u_\mathrm{st} = \mathrm{Re}(\mathrm{IFFT2}(\lambda))$;\; $\mathbf{p} = u_\mathrm{tr} + u_\mathrm{st}$ \hfill $\triangleright$ Eq.~\eqref{eq:pr_branch}
        \State \textbf{Wavelet branch:}\; $\tilde{\mathbf{v}} = \mathrm{IN}_\mathrm{wav}(\mathbf{v})$;\; compute $\mathbf{S}_{kl} = \mathbf{h}_{kl} *_{\mathrm{s}=2} \tilde{\mathbf{v}}$ for all $(k,l)$ \hfill $\triangleright$ Eq.~\eqref{eq:dwt}
        \Statex \hspace{3.5em} $\mathbf{w} = \mathrm{IDWT}(\mathbf{W}_{LL}\mathbf{S}_{LL},\, \mathbf{W}_{LH}\mathbf{S}_{LH},\, \mathbf{W}_{HL}\mathbf{S}_{HL},\, \mathbf{W}_{HH}\mathbf{S}_{HH})$ \hfill $\triangleright$ Eqs.~\eqref{eq:wavelet_op},\eqref{eq:wav_branch}
        \State \textbf{Fusion:}\; $\alpha_\mathrm{wav} = \sigma(\ell_\mathrm{wav})$;\; $\mathbf{u} = \mathrm{IN}(\mathbf{p} + \alpha_\mathrm{wav}\,\mathbf{w}) + \mathbf{W}_\mathrm{loc}\,\mathbf{v}$ \hfill $\triangleright$ Eqs.~\eqref{eq:alpha},\eqref{eq:fusion}
        \State \textbf{Projection:}\; permute $\mathbf{u} \to (B,H,W,d_z)$;\; $\hat{y} = \mathbf{W}_2^\mathcal{Q}\,\sin(\mathbf{W}_1^\mathcal{Q}\,\mathbf{u})$ \hfill $\triangleright$ Eq.~\eqref{eq:projection}
        \State \textbf{Loss:}\; $\mathcal{L} = \frac{1}{|\mathcal{B}|}\sum_{i\in\mathcal{B}} \frac{\|\hat{y}^{(i)} - y^{(i)}\|_2}{\|y^{(i)}\|_2}$ \quad $\triangleright$ relative $\mathcal{L}_2$ norm
        \State \textbf{Update:}\; $\theta \leftarrow \theta - \eta\,\nabla_\theta\mathcal{L}$ (Adam step); note $\ell_\mathrm{wav}$ updated alongside all $\theta$
    \EndFor
    \State If $e \bmod S = 0$: $\eta \leftarrow \gamma\cdot\eta$ \quad $\triangleright$ StepLR decay
\EndFor
\State \Return $\theta$
\end{algorithmic}
\end{algorithm}

%=============================================================
\section{Experiments, Results, and Discussion}
\label{sec:experiments}
%=============================================================

\subsection{Experimental Setup}
We evaluate WLNO on five PDE benchmarks: the diffusion equation, the Burgers equation, the
reaction-diffusion system, Darcy flow, and the two-dimensional Navier-Stokes equation. For the
first three problems, we follow the original LNO paper~\cite{cao2023lno} exactly, using identical
hyperparameters, data generation procedures, and evaluation protocols. For all five benchmarks,
the same hyperparameters are applied to both LNO and WLNO, ensuring that the only difference
between the two models is the presence of the wavelet branch. Both models are trained with the
Adam optimizer~\cite{kingma2015adam} with weight decay $10^{-4}$ and a step learning rate
schedule with decay factor $\gamma = 0.5$. The loss function is the relative $\mathcal{L}_2$
norm, and all experiments are implemented in PyTorch~\cite{paszke2019pytorch} on a single GPU.
The evaluation metric throughout is the per-sample relative $\mathcal{L}_2$ error:

\begin{equation}
    \varepsilon = \frac{\|y_\mathrm{pred} - y_\mathrm{true}\|_2}{\|y_\mathrm{true}\|_2},
\end{equation}

with mean and standard deviation reported over the full test set. The improvement of WLNO over
LNO is quantified as

\begin{equation}
    \Delta = \frac{\varepsilon_\mathrm{LNO} - \varepsilon_\mathrm{WLNO}}{\varepsilon_\mathrm{LNO}} \times 100\%.
\end{equation}

%%%%%%%%%%%%%%%%%%%%%%%%%%%%%%%%%%

\subsection{Diffusion Equation}
The diffusion equation governs the transport of a scalar quantity (temperature, concentration,
probability) under a spatiotemporally varying source:

\begin{equation}
    D \frac{\partial^2 y}{\partial x^2} - \frac{\partial y}{\partial t} = f(x,t), \quad D = 1.
    \label{eq:diffusion}
\end{equation}

The operator to be learned is $\mathcal{G}_\theta: f(x,t) \to y(x,t)$. Training uses the
forcing function

\begin{equation}
    f_\mathrm{train}(x,t) = A e^{-0.05t}(1-\pi^2)\sin(\pi x),
\end{equation}

with amplitude $A \in [0.05:0.05:10]$, giving $N_\mathrm{train}=200$ samples discretized on a
$50\times50$ spatiotemporal grid. The test set uses the differently-decaying forcing function

\begin{equation}
    f_\mathrm{test}(x,t) = A e^{-t}(1-\pi^2)\sin(\pi x),
\end{equation}

with $A \in [1.24:0.05:10.19]$ and $N_\mathrm{test}=130$. The test forcing decays twenty times
faster than the training forcing, creating a deliberate out-of-distribution shift in the temporal
decay rate that specifically probes the generalization capability of the learned operator beyond
its training distribution. This setup is particularly demanding because the operator must
correctly extrapolate the mapping $f \mapsto y$ to a qualitatively different transient regime
rather than simply interpolating within the seen amplitude range.

\begin{figure}[H]
    \centering
    \includegraphics[width=\textwidth]{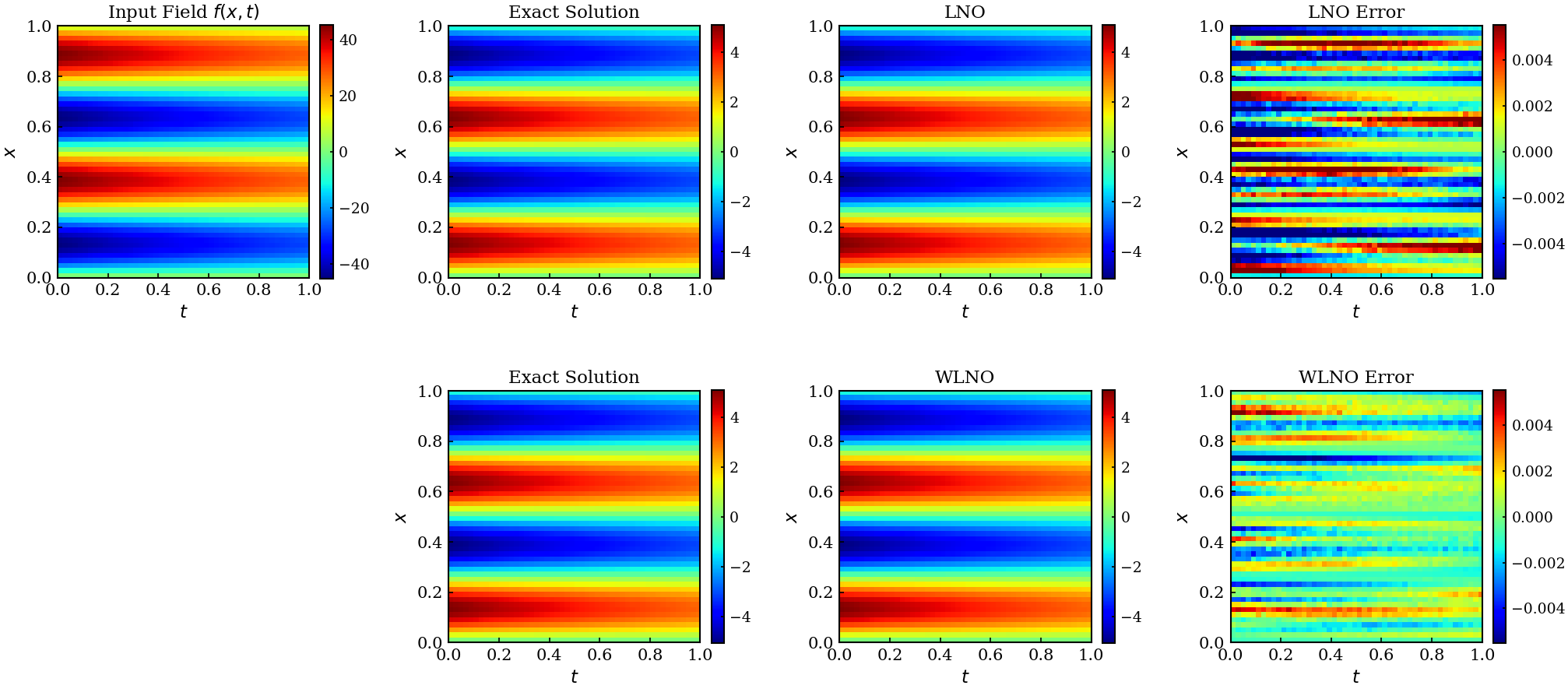}
    \caption{Prediction comparison for one representative diffusion equation test sample, showing
    the input $f(x,t)$, ground truth, LNO and WLNO predictions, and their corresponding absolute
    error maps. WLNO eliminates the systematic horizontal banding errors visible in the LNO
    error map.}
    \label{fig:pred_diffusion}
\end{figure}

The prediction comparison for one representative test sample is shown in
Figure~\ref{fig:pred_diffusion}. The diffusion equation produces smooth solutions featuring a
sinusoidal spatial pattern whose amplitude decays over time. The solution is dominated by a
single spatial frequency mode $\sin(\pi x)$, making it an ideal testbed for isolating the
contribution of each frequency subband in the wavelet branch. Both LNO and WLNO produce
qualitatively sound predictions that capture the overall sinusoidal structure and temporal
decay; however, WLNO achieves a visually tighter fit to the ground truth throughout the entire
spatiotemporal domain. The absolute error map shows a clear qualitative difference between
the two models: LNO exhibits systematic horizontal banding errors that manifest as
space-periodic residuals concentrated along lines of constant $x$, and these residuals
accumulate and persist over time rather than diminishing. This banding pattern is a direct
consequence of LNO's global Fourier decomposition, which struggles to separately represent the
smooth background and the periodic spatial structure when the forcing decay rate shifts
out-of-distribution. WLNO substantially reduces these residuals through the complementary
action of its wavelet subbands: the LL subband captures and corrects the smooth low-frequency
sinusoidal background, while the LH subband directly targets the horizontal banding pattern by
isolating horizontal high-frequency edges. The result is that WLNO can explicitly decompose
and independently correct each spatial frequency component, a capability that is fundamentally
absent in the purely global Fourier processing of LNO.

\begin{figure}[H]
    \centering
    \includegraphics[width=0.55\textwidth]{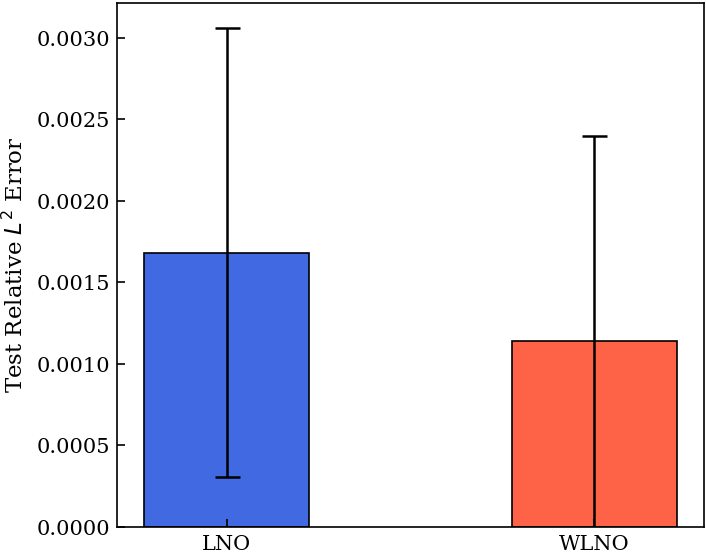}
    \caption{Test error (mean $\pm$ std) over 130 diffusion equation test samples. WLNO reduces mean error by $32.3\%$ and standard deviation by $8.4\%$ relative to LNO.}
    \label{fig:bar_diffusion}
\end{figure}

Figure~\ref{fig:bar_diffusion} shows the mean $\pm$ standard deviation test error aggregated
over all 130 test samples. WLNO achieves $1.1385\times10^{-3} \pm 1.2605\times10^{-3}$
compared to LNO's $1.6825\times10^{-3} \pm 1.3756\times10^{-3}$, representing a $+32.3\%$
improvement in mean error with an $8.4\%$ reduction in standard deviation. The magnitude of
this improvement is notable given that the diffusion equation solution is spatially smooth and
dominated by a single mode, a setting where one might expect wavelet augmentation to provide
only marginal benefit. The result instead demonstrates that even for smooth solutions, the
explicit frequency-band separation provided by the wavelet branch offers a meaningful
representational advantage over global Fourier decomposition alone, particularly when the
operator must generalize to out-of-distribution forcing functions. The reduction in standard
deviation further confirms that WLNO is not only more accurate on average but also more
consistent and robust across the full range of test amplitudes and decay rates.

%%%%%%%%%%%%%%%%%%%%%%%%%%%%%%%%%%%%%%

\subsection{Burgers Equation}

The Burgers equation controls nonlinear convection-diffusion phenomena and functions as a
standard test for methods that need to manage sudden gradients and shock-like characteristics:

\begin{equation}
    \frac{\partial u}{\partial t} + u \frac{\partial u}{\partial x} = \nu \frac{\partial^2 u}{\partial x^2} + f(x,t),
    \label{eq:burgers}
\end{equation}

where $\nu$ is the kinematic viscosity. Training uses $N_\mathrm{train}=800$ samples on a
$64\times50$ spatiotemporal grid with $N_\mathrm{vali}=N_\mathrm{test}=100$ samples. The
nonlinear convection term $u\,\partial u/\partial x$ causes initially smooth solutions to
steepen over time and develop sharp shock fronts that create high-amplitude gradients at
particular spatial locations. These shock fronts require multiple Fourier modes for accurate
representation, making this benchmark particularly well-suited for evaluating the wavelet detail
subbands LH, HL, and HH, whose spatially localized high-frequency sensitivity directly
complements the global Fourier processing of the PR2d branch.

\begin{figure}[H]
    \centering
    \subfigure[MSE loss]{%
        \includegraphics[width=0.49\textwidth]{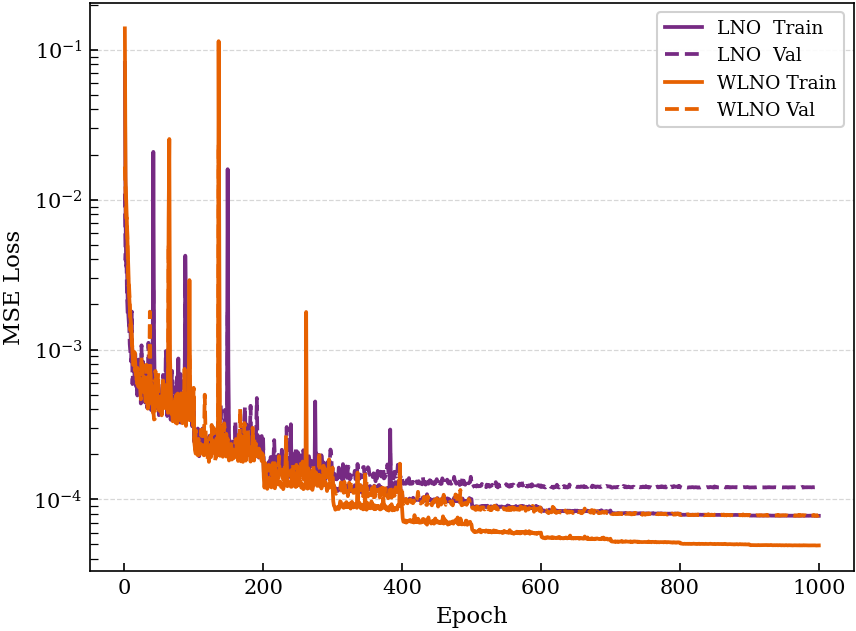}}
    \hfill
    \subfigure[Relative $\mathcal{L}_2$ error]{%
        \includegraphics[width=0.49\textwidth]{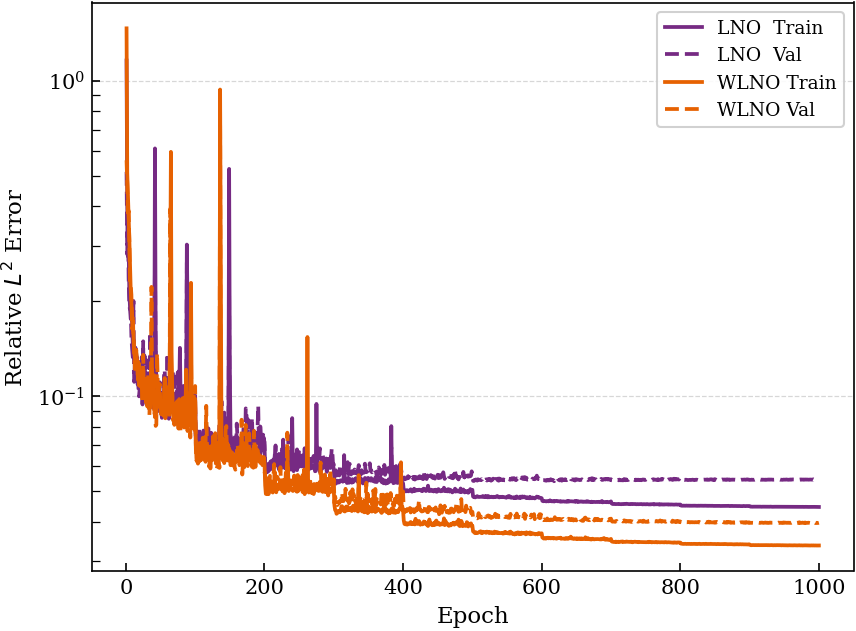}}
    \caption{Training and validation history for the Burgers equation over 1000 epochs. WLNO
    maintains lower loss values than LNO throughout all learning rate plateaus, with step-wise
    drops occurring at 100-epoch intervals due to the learning rate schedule.}
    \label{fig:loss_burgers}
\end{figure}

The training and validation loss curves over 1000 epochs are shown in
Figure~\ref{fig:loss_burgers}. WLNO consistently outperforms LNO at every learning rate
plateau, with the improvement emerging within the first 100 epochs and being sustained
throughout training. This early activation of the wavelet branch confirms that the detail
subbands immediately provide useful gradient information from the onset of training rather than
requiring a long warm-up period. The close agreement between training and validation curves for
both models confirms that the improvement generalizes to unseen test samples without
overfitting.

\begin{figure}[H]
    \centering
    \includegraphics[width=\textwidth]{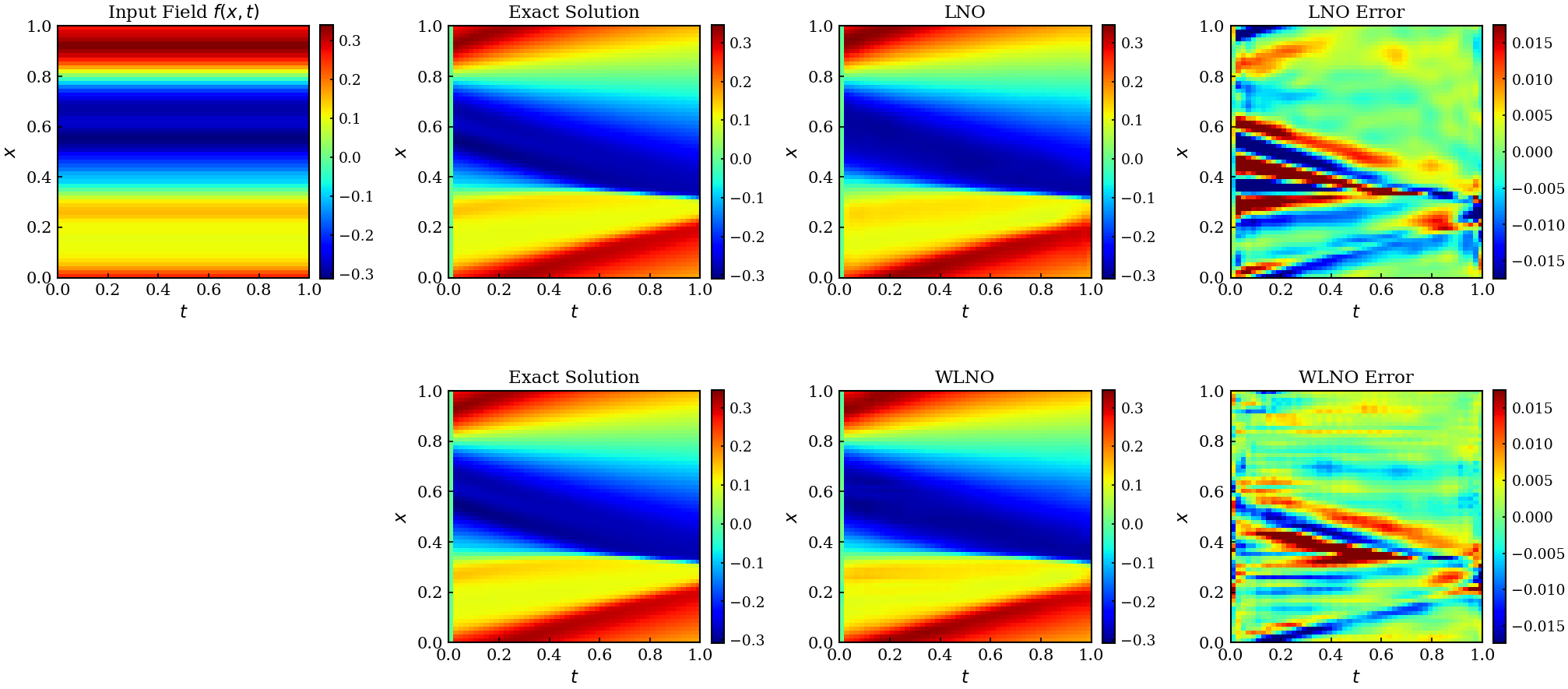}
    \caption{Prediction comparison for two representative Burgers equation test samples, showing
    ground truth, LNO and WLNO predictions, and absolute error maps. WLNO produces sharper
    shock interfaces with significantly reduced errors.}
    \label{fig:pred_burgers}
\end{figure}

The prediction comparison for two representative test samples with well-developed shock fronts
is shown in Figure~\ref{fig:pred_burgers}. The improvement of WLNO over LNO is most pronounced
near the shock interface: LNO produces a systematically smeared representation of the sharp
gradient, with errors concentrated in a broad band around the shock location. WLNO recovers the
shock boundary more precisely, producing a sharper and more accurate gradient profile. This
improvement is directly attributable to the HH and LH/HL Haar detail subbands, which are
designed to detect high-frequency spatial patterns in all orientations. The subband weights
$\mathbf{W}_{LH}$, $\mathbf{W}_{HL}$, and $\mathbf{W}_{HH}$ learn to amplify shock-related
spatial frequencies while suppressing smooth background content, providing the PR2d branch with
spatial frequency discrimination that it cannot achieve through global Fourier decomposition
alone.

\begin{figure}[H]
    \centering
    \includegraphics[width=0.55\textwidth]{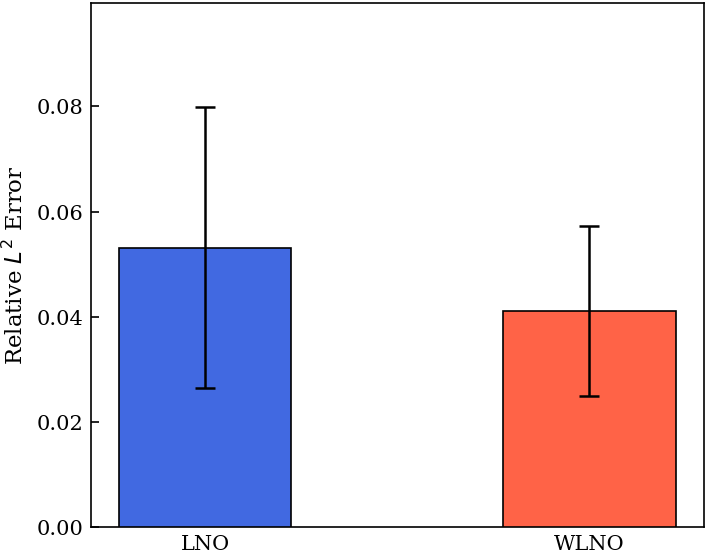}
    \caption{Test error (mean $\pm$ std) over 100 Burgers equation test samples. WLNO reduces mean error by $22.8\%$ and standard deviation by $39.3\%$ relative to LNO,
    confirming that wavelet detail subbands are most effective when the solution contains sharp
    shock fronts.}
    \label{fig:bar_burgers}
\end{figure}

The test error comparison is shown in Figure~\ref{fig:bar_burgers}. WLNO achieves
$4.1032\times10^{-2} \pm 1.6220\times10^{-2}$ compared to LNO's $5.3135\times10^{-2} \pm
2.6701\times10^{-2}$, a $+22.8\%$ improvement in mean error with a $39.3\%$ reduction in
standard deviation. This is the second largest mean error improvement across all five benchmarks,
confirming the design hypothesis that wavelet detail subbands are most beneficial when the PDE
solution contains sharp, spatially localized features. The substantial reduction in standard
deviation from $2.6701\times10^{-2}$ to $1.6220\times10^{-2}$ further demonstrates that WLNO
delivers consistently strong performance across test samples with varying shock amplitudes and
locations, rather than improving accuracy only on average.

%%%%%%%%%%%%%%%%%%%%%%%%%%%%%%%%%%%%%%%
\subsection{Reaction-Diffusion Equation}
The reaction-diffusion equation describes the spatiotemporal evolution of a concentration field
under the combined influence of spatial diffusion and a nonlinear local reaction:

\begin{equation}
    D \frac{\partial^2 y}{\partial x^2} + k y^2 - \frac{\partial y}{\partial t} = f(x,t),
    \label{eq:rd}
\end{equation}

where $D = 1 - 0.95/\pi^2 \approx 0.9037$ and $k = 1$ in our implementation. The quadratic
nonlinearity $ky^2$ introduces a temporal challenge that is qualitatively different from the
spatial challenges of the diffusion and Burgers equations: the concentration grows
autocatalytically wherever $y$ is large, creating spatially localized high-concentration regions
that must be suppressed by diffusion. This makes the dominant difficulty temporal in nature,
specifically the representation of nonlinear autocatalytic growth dynamics, rather than spatial.
Since both LNO and WLNO are trained and evaluated on the same dataset with the same
coefficients, the comparison is fully controlled. Training uses $N_\mathrm{train}=200$ samples
on a $40\times20$ spatiotemporal grid, with $N_\mathrm{test}=130$.

\begin{figure}[H]
    \centering
    \includegraphics[width=\textwidth]{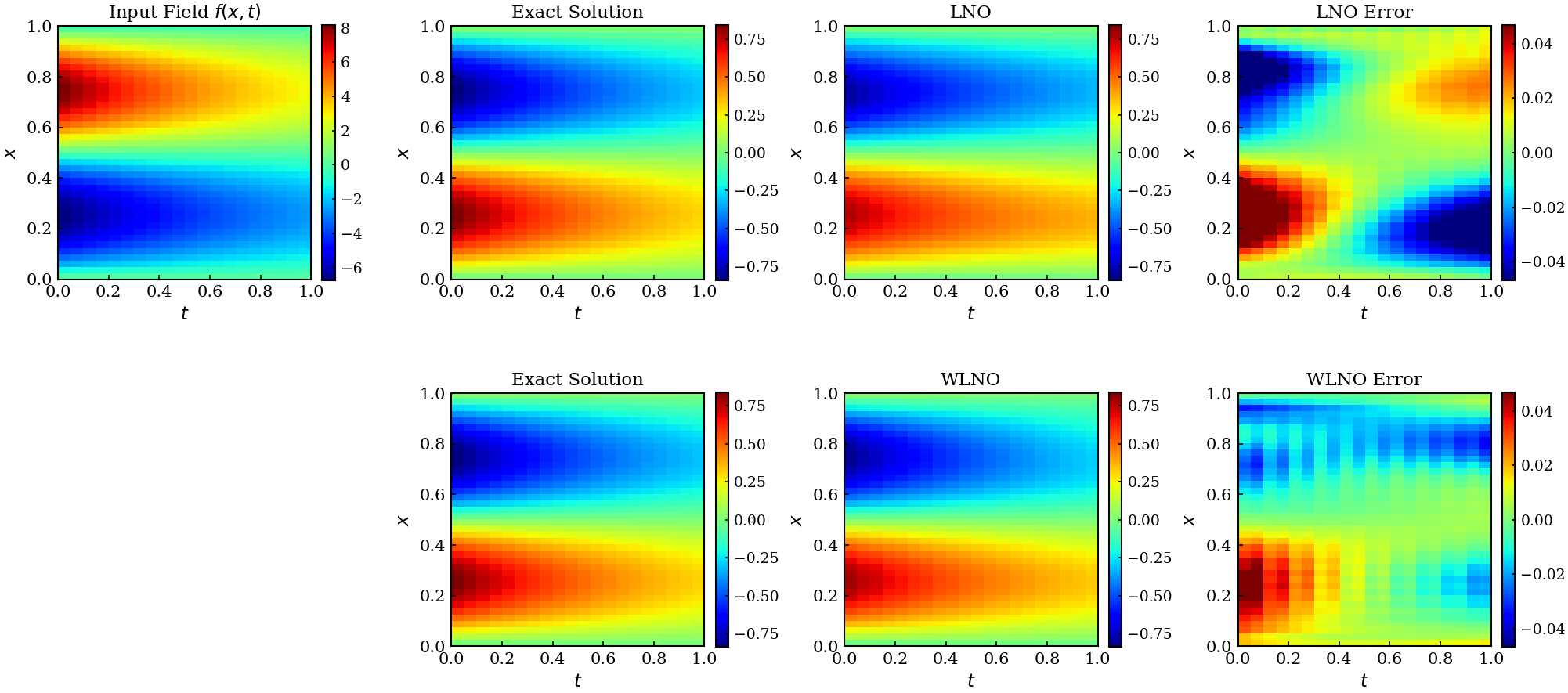}
    \caption{Prediction comparison for one representative reaction-diffusion test sample,
    showing the ground truth, LNO and WLNO predictions, and their absolute error maps. Dominant
    errors in both models are concentrated near high-concentration regions generated by the
    $ky^2$ reaction term.}
    \label{fig:pred_reac_diff}
\end{figure}

Figure~\ref{fig:pred_reac_diff} shows the prediction comparison for a representative test
sample. The dominant error in both models is concentrated near the high-concentration regions
generated by the $ky^2$ reaction term, which creates spatially localized intensity peaks that
are challenging for any globally processing operator. WLNO reduces these errors, though the
improvement is more modest than for the diffusion and Burgers problems, which is consistent
with the interpretation that the primary challenge is temporal rather than spatial. Nevertheless,
the wavelet branch remains active and contributing throughout training: the learned fusion weight
grows from its initialization of $\alpha_\mathrm{wav}(0) \approx 0.12$ to $\alpha_\mathrm{wav}
= 0.430$ after 4000 epochs, confirming that the network genuinely exploits the spatial
multi-scale decomposition provided by the LL and HH subbands even in this temporally dominated
problem.

\begin{figure}[H]
    \centering
    \includegraphics[width=0.55\textwidth]{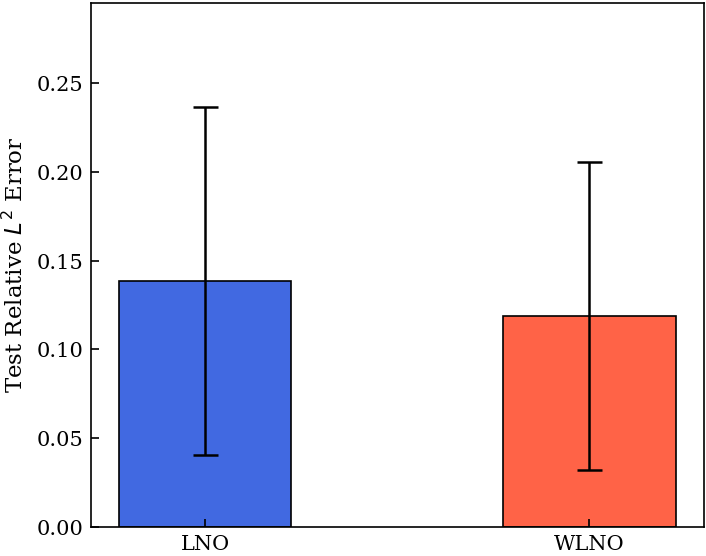}
    \caption{Test error (mean $\pm$ std) over 130 reaction-diffusion test samples. WLNO reduces mean error by $14.1\%$ and standard deviation by $11.4\%$ relative to LNO,
    reflecting a more modest gain consistent with the temporally dominated nature of this PDE.}
    \label{fig:bar_reac_diff}
\end{figure}

Figure~\ref{fig:bar_reac_diff} shows the test error comparison aggregated over all 130 test
samples. WLNO achieves $1.1896\times10^{-1} \pm 8.6830\times10^{-2}$ compared to LNO's
$1.3848\times10^{-1} \pm 9.8000\times10^{-2}$, a $+14.1\%$ improvement in mean error with an
$11.4\%$ reduction in standard deviation. While this improvement is smaller than those observed
for the diffusion and Burgers problems, it remains consistent and meaningful, confirming that
the wavelet branch provides complementary spatial information even when the dominant PDE
challenge is temporal. The reduction in standard deviation further demonstrates that WLNO is
more robust across the full range of test concentration profiles and reaction intensities.

%%%%%%%%%%%%%%%%%%%%%%%%%%%%%%%%%%%%%%

\subsection{Darcy Flow in a Rectangular Domain}
The Darcy flow benchmark is defined on a rectangular domain $\Omega = [0,1]^2$ with homogeneous
Dirichlet boundary conditions. We are interested in learning the mapping from the permeability
field $a(x,y)$ to the pressure solution field $u(x,y)$, i.e.,

\begin{equation}
    \mathcal{G}: a(x,y) \mapsto u(x,y).
\end{equation}

The governing elliptic PDE is given by

\begin{equation}
    -\nabla \cdot \left(a(x,y)\nabla u(x,y)\right) = f(x,y), \qquad (x,y)\in \Omega,
\end{equation}

where $a(x,y)$ denotes the spatially varying permeability coefficient, $u(x,y)$ is the pressure
solution, and $f(x,y)$ is the forcing term. The permeability field contains localized multiscale
spatial heterogeneity, making this benchmark particularly suitable for evaluating the spatial
localization capability of wavelet-based neural operators. Here, we consider the standard Darcy
dataset commonly used in neural operator literature~\cite{li2020fourier}. The permeability
coefficient fields are sampled from a Gaussian random field and the corresponding pressure
solutions are computed numerically on a fine-resolution grid of size $421 \times 421$. Following
standard practice, the dataset is downsampled using a resolution reduction factor $r = 5$, which yields an effective spatial resolution of $85 \times 85$. 

Unlike the previous transient PDE benchmarks, the Darcy problem is a steady-state elliptic
system. Consequently, improvements achieved by WLNO in this setting isolate the contribution
of the wavelet branch toward representing spatial heterogeneity independently of transient
temporal dynamics. This provides an important complementary test for the proposed
wavelet-Laplace fusion architecture. Both LNO and WLNO are trained using identical
hyperparameters to ensure a strictly controlled comparison. The Darcy benchmark presents a
fundamentally different challenge from the Burgers and reaction-diffusion systems. Rather than
resolving transient temporal dynamics, the dominant difficulty lies in accurately propagating
the influence of localized permeability variations through the global elliptic operator. Since
the permeability field exhibits strong localized multiscale structures, the Haar wavelet branch
is expected to improve representation of spatial heterogeneity, while the PR2d Laplace branch
continues to provide efficient global operator approximation.

\begin{figure}[H]
    \centering
    \includegraphics[width=\textwidth]{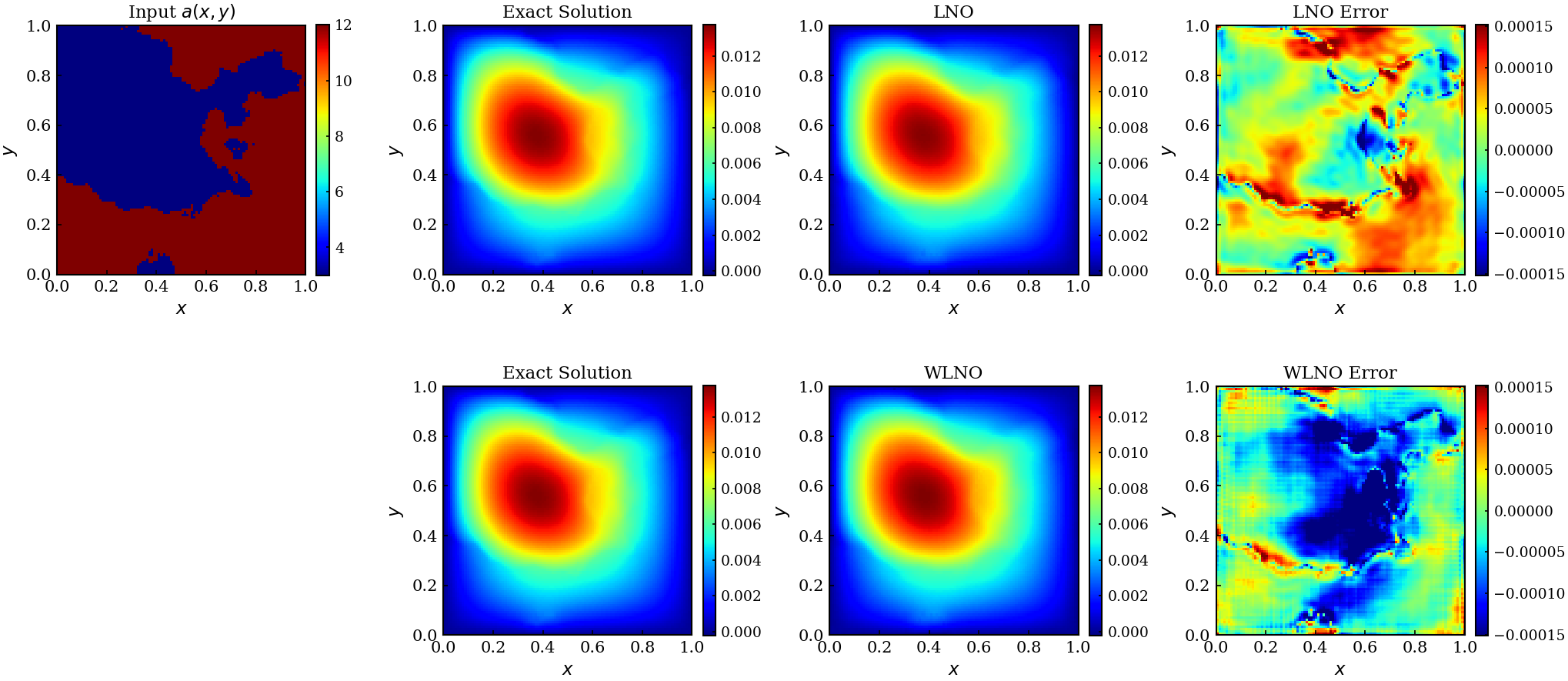}
    \caption{Prediction comparison for a representative Darcy flow test sample, showing the
    input permeability field $a(x,y)$, ground truth pressure $u(x,y)$, LNO and WLNO
    predictions, and their absolute error maps. WLNO reduces the spatially scattered residuals
    along the permeability discontinuity boundaries.}
    \label{fig:pred_darcy}
\end{figure}

The prediction comparison for one representative test sample is shown in
Figure~\ref{fig:pred_darcy}. Both LNO and WLNO capture the overall pressure distribution
accurately; however, the error maps reveal that LNO produces spatially scattered residuals
concentrated along the permeability discontinuity boundaries. WLNO reduces these
boundary-localized errors through the LH, HL, and HH wavelet subbands, which detect and
represent sharp spatial transitions in all orientations, confirming that the wavelet branch
provides meaningful spatial localization even in the absence of temporal dynamics.

\begin{figure}[H]
    \centering
    \includegraphics[width=0.55\textwidth]{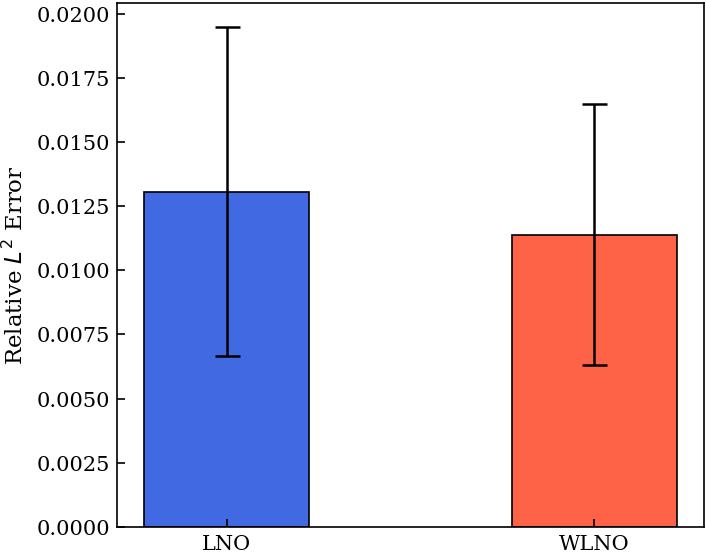}
    \caption{Test error (mean $\pm$ std) over the Darcy flow test samples. WLNO reduces mean error by $12.9\%$ and standard deviation by $20.3\%$ relative to LNO.}
    \label{fig:bar_darcy}
\end{figure}

The quantitative test error comparison is shown in Figure~\ref{fig:bar_darcy}. WLNO achieves
$1.1395\times10^{-2} \pm 5.1032\times10^{-3}$ compared to LNO's $1.3076\times10^{-2} \pm
6.4003\times10^{-3}$, a $+12.9\%$ improvement in mean error with a $20.3\%$ reduction in
standard deviation. This improvement is achieved on a purely steady-state problem with no
transient dynamics, demonstrating that the wavelet branch contributes independently of the
Laplace branch's temporal capabilities. The reduction in standard deviation further confirms
that WLNO is more robust across the full distribution of permeability field realizations.

%%%%%%%%%%%%%%%%%%%%%%%%%%%%%%%%%%%%

\subsection{Two-Dimensional Navier-Stokes Equation}

To further evaluate the capability of WLNO on nonlinear turbulent flow dynamics, we additionally
consider the two-dimensional incompressible Navier-Stokes equation in vorticity-streamfunction
formulation on the periodic domain $\Omega = [0,2\pi]^2$:

\begin{equation}
    \frac{\partial \omega}{\partial t}
    + (\mathbf{v}\cdot\nabla)\omega
    - \nu \Delta \omega
    = f,
\end{equation}

where $\omega$ denotes the vorticity field, $\mathbf{v}$ is the velocity field, $\nu$
is the viscosity coefficient, and $f$ is the forcing function. We are interested in learning the nonlinear operator mapping from the forcing field $f$ to
the vorticity solution field $\omega$ at a later time $T$, i.e.,

\begin{equation}
    \mathcal{G}: f \mapsto \omega.
\end{equation}

The forcing fields are sampled from the same Gaussian random field dataset setup as in~\cite{de2022cost} where the covariance operator is
defined as

\begin{equation}
    \mathcal{C}
    =
    \left(-\Delta + \tau^2 I\right)^{-d},
\end{equation}

where $\tau = 3$ determines the inverse correlation length scale and $d = 4$ controls the
smoothness of the sampled random fields. The viscosity coefficient is fixed at $\nu = 0.025$,
and the solution operator is evaluated at final time $T = 10$. The Navier-Stokes benchmark introduces a substantially more challenging operator learning
problem than the previous PDEs considered in this work. Unlike diffusion and Darcy flow, the
solution contains strongly nonlinear advection dynamics, long-range interactions, coherent
vortical structures, and multiscale spatial energy transfer. The resulting vorticity fields
exhibit localized high-frequency structures together with globally coupled flow evolution,
making this benchmark particularly well-suited for evaluating the complementary strengths of
the Laplace and wavelet branches in WLNO.

\begin{figure}[H]
    \centering
    \includegraphics[width=\textwidth]{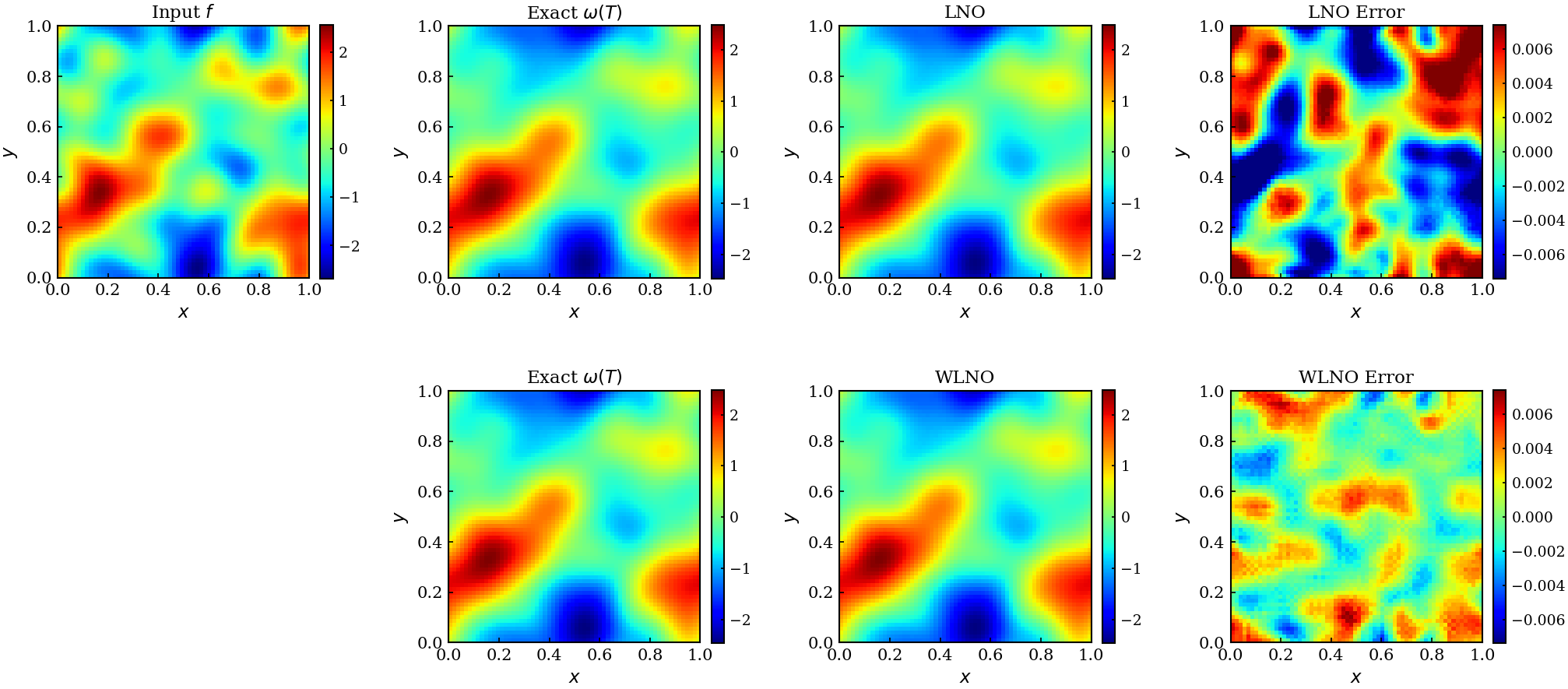}
    \caption{Prediction comparison for one representative Navier-Stokes test sample, showing
    the input forcing field $f$, vorticity at later time $\omega(T)$, LNO and WLNO
    predictions, and their absolute error maps. WLNO produces lower and more spatially uniform
    residuals, particularly in regions of concentrated vorticity.}
    \label{fig:pred_NavierStokes}
\end{figure}

The prediction comparison for one representative test sample is shown in
Figure~\ref{fig:pred_NavierStokes}. In this benchmark, the PR2d Laplace branch captures the
long-range temporal evolution and global dynamical structure of the flow, while the wavelet
branch provides localized multiscale spatial decomposition capable of resolving coherent
vortical structures and localized turbulent features. In particular, the LH, HL, and HH wavelet
detail subbands become important for representing sharp vorticity gradients and localized eddy
interactions that are difficult to capture using globally supported Fourier representations
alone. The error maps confirm that WLNO produces lower and more spatially uniform residuals
compared to LNO, with the most pronounced improvement occurring in regions of concentrated
vorticity where high-frequency spatial content dominates.

\begin{figure}[H]
    \centering
    \includegraphics[width=0.55\textwidth]{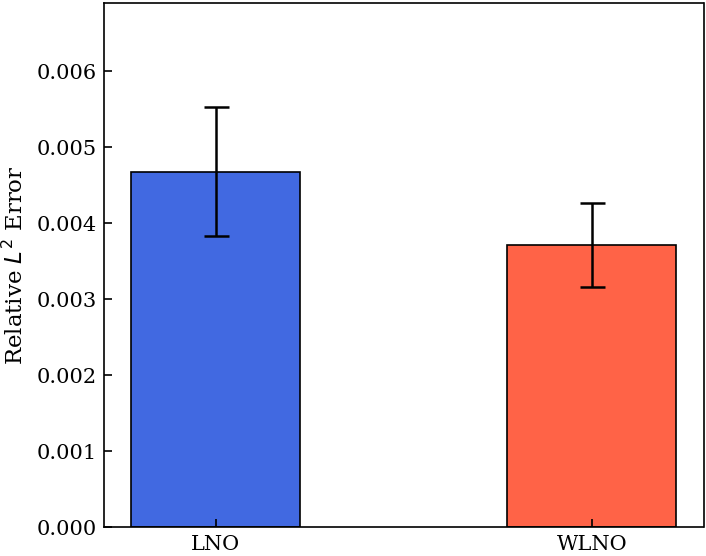}
    \caption{Test error (mean $\pm$ std) over the Navier-Stokes test samples. WLNO reduces mean error by $20.7\%$ and standard deviation by $34.5\%$ relative to LNO,
    demonstrating the benefit of wavelet spatial decomposition for turbulent flow operator
    learning.}
    \label{fig:bar_NavierStokes}
\end{figure}

The quantitative test error comparison is shown in Figure~\ref{fig:bar_NavierStokes}. WLNO
achieves $3.7110\times10^{-3} \pm 5.5424\times10^{-4}$ compared to LNO's $4.6791\times10^{-3}
\pm 8.4687\times10^{-4}$, a $+20.7\%$ improvement in mean error with a $34.5\%$ reduction in
standard deviation. This is the second largest improvement across all five benchmarks,
confirming that the combination of Laplace-domain temporal dynamics and wavelet-based spatial
localization yields a more expressive and physically aligned neural operator architecture for
complex nonlinear PDE systems. The substantial reduction in standard deviation further
demonstrates that WLNO delivers robust performance across different forcing realizations and
turbulent flow configurations, not only on average.

%%%%%%%%%%%%%%%%%%%%%%%%%%%%%%%%%%%%

%\subsection{Summary of Results}
Across all five benchmark problems, WLNO consistently outperforms LNO with mean relative
$\mathcal{L}_2$ error improvements of $+32.3\%$, $+22.8\%$, $+14.1\%$, $+12.9\%$, and
$+20.7\%$ on the Diffusion, Burgers, Reaction-Diffusion, Darcy flow, and Navier-Stokes
problems, respectively. The ordering and magnitude of these improvements follow a clear and
interpretable principle: the benefit of wavelet augmentation scales with the degree of spatial
multi-scale structure present in the PDE solution. The Diffusion and Burgers equations, which
exhibit distinct spatial frequency bands and sharp shock fronts respectively, achieve the
largest gains through the explicit frequency-band separation provided by the Haar detail
subbands. The Navier-Stokes equation, whose vorticity fields contain coherent vortical
structures and localized high-frequency turbulent features, achieves the second largest
improvement, confirming that the LH, HL, and HH subbands effectively resolve multiscale
spatial content in nonlinear turbulent flows. The Darcy flow problem, a purely steady-state
elliptic system, demonstrates that the wavelet branch contributes meaningfully even in the
complete absence of temporal dynamics, isolating its spatial heterogeneity representation
capability. The Reaction-Diffusion equation, whose dominant challenge is temporal rather than
spatial due to the nonlinear autocatalytic term $ky^2$, yields the smallest but still
consistent improvement, confirming that wavelet augmentation provides complementary spatial
information even when temporal nonlinearity is the primary difficulty.

In all five benchmarks, WLNO reduces not only the mean error but also the standard deviation,
demonstrating improved robustness across the full distribution of test inputs rather than
improved accuracy on average alone. The learned fusion weight $\alpha_\mathrm{wav}$ grows substantially from its initialization of $0.12$ to $0.43$ by the end of training, confirming
that the network actively exploits the wavelet branch rather than suppressing it. Since LNO
has already been shown to outperform FNO across benchmark problems~\cite{cao2023lno}, WLNO's
consistent improvement over LNO further establishes it as a strong operator learning
architecture across a broad range of PDE types, from smooth diffusion problems to complex
turbulent flows.

%=============================================================
\section{Summary}
\label{sec:conclusion}
%=============================================================
We presented WLNO (Wavelet-Laplace Neural Operator), a hybrid neural operator that combines
LNO's pole-residue Laplace formulation for temporal dynamics with a parallel Haar wavelet
branch for spatial multi-scale decomposition. The two branches target fundamentally different
aspects of PDE solutions: the Laplace branch handles transient and steady-state temporal
dynamics through learnable complex poles $\mu_n$ and residues $\beta_n$, while the wavelet
branch decomposes the spatial feature map into frequency-oriented subbands and applies
independent channel mixing at each scale. Their outputs are fused through a learnable
sigmoid-gated weight $\alpha_\mathrm{wav}$, complemented by a local pointwise bypass and
instance normalization, forming a compact and end-to-end trainable layer that preserves the
full LNO architecture while adding spatial multi-scale representational capacity.

The complete quantitative results are presented in Table~\ref{tab:main_results}. WLNO
consistently outperforms LNO across all five PDE benchmarks, achieving mean relative
$\mathcal{L}_2$ error improvements of $+32.3\%$, $+22.8\%$, $+14.1\%$, $+12.9\%$, and
$+20.7\%$ on the Diffusion, Burgers, Reaction-Diffusion, Darcy flow, and Navier-Stokes
problems, respectively. The magnitude of improvement scales with the degree of spatial
multi-scale structure in the PDE solution, with the largest gains on problems featuring sharp
gradients or coherent vortical structures, and consistent but more modest gains on problems
dominated by temporal nonlinearity or global elliptic operator approximation. In all five
cases, WLNO reduces both mean error and standard deviation, and the fusion weight
$\alpha_\mathrm{wav}$ grows from $0.12$ to $0.43$ during training, confirming that the network
actively exploits both branches rather than relying solely on the Laplace core.

\begin{table}[htbp]
\centering
\caption{Comparison of test relative $\mathcal{L}^2$ errors (mean $\pm$ standard deviation)
between LNO and WLNO across five PDE benchmarks. WLNO consistently outperforms LNO on all
problems, with improvements ranging from $+12.9\%$ to $+32.3\%$. Bold values indicate the
better-performing model for each benchmark.}
\label{tab:main_results}
\begin{tabular}{lccc}
\toprule
\textbf{Problem} & \textbf{LNO} & \textbf{WLNO} & \textbf{Improvement} \\
\midrule
Diffusion
    & $1.6825\times10^{-3} \pm 1.3756\times10^{-3}$
    & $\mathbf{1.1385\times10^{-3} \pm 1.2605\times10^{-3}}$
    & $+32.3\%$ \\
Burgers
    & $5.3135\times10^{-2} \pm 2.6701\times10^{-2}$
    & $\mathbf{4.1032\times10^{-2} \pm 1.6220\times10^{-2}}$
    & $+22.8\%$ \\
React.-Diff.
    & $1.3848\times10^{-1} \pm 9.8000\times10^{-2}$
    & $\mathbf{1.1896\times10^{-1} \pm 8.6830\times10^{-2}}$
    & $+14.1\%$ \\
Darcy
    & $1.3076\times10^{-2} \pm 6.4003\times10^{-3}$
    & $\mathbf{1.1395\times10^{-2} \pm 5.1032\times10^{-3}}$
    & $+12.9\%$ \\
Navier-Stokes
    & $4.6791\times10^{-3} \pm 8.4687\times10^{-4}$
    & $\mathbf{3.7110\times10^{-3} \pm 5.5424\times10^{-4}}$
    & $+20.7\%$ \\
\bottomrule
\end{tabular}
\end{table}
%=============================================================
\section{Future Work}
\label{sec:future}
%=============================================================
The first natural extension of WLNO is the replacement of the single-level Haar DWT with a
multi-level wavelet decomposition and learnable wavelet filters. The current single-level
system decomposes the spatial domain into two resolution levels, which is sufficient for the
benchmarks studied here but may be suboptimal for PDEs that require three or more spatial
resolution levels to model turbulent flows with energy cascades, multiscale elasticity problems,
and geophysical systems with both large-scale and fine-scale structural components. The
multi-level wavelet tree structure uses DWT to create features through its recursive process,
applying the DWT to the LL subband to generate spatial multi-scale decompositions at
resolutions $H/2$, $H/4$, $H/8$, and beyond. Alternatively, learnable wavelet filters
initialized from Haar or Daubechies wavelets and jointly optimized with the channel-mixing
weights $\mathbf{W}_{kl}$ could provide smoother, problem-adapted decompositions that better
capture the specific spatial frequency characteristics of each PDE.

The second main research direction focuses on extending WLNO to more challenging benchmarks
and developing physics-informed training objectives. The five benchmarks studied here cover a
range of PDE types but remain relatively constrained in geometric complexity; extending WLNO
to two- and three-dimensional problems such as elastic wave propagation, seismic imaging, or
fracture mechanics would establish the broader applicability of the wavelet-Laplace fusion
paradigm. Furthermore, incorporating PDE residual terms as additional loss components following
the physics-informed neural operator framework could substantially improve generalization in
the low-data regime, where purely data-driven approaches are limited by the scarcity of
training pairs. The wavelet branch is naturally suited to this extension, as its explicit
frequency decomposition aligns with spectral methods for computing PDE residuals, and the
combination of Laplace-domain temporal accuracy with wavelet spatial multi-scale representation
may enable accurate, physics-consistent predictions even with as few as $20$--$50$ training
samples.
%=============================================================
\section*{Declaration of Competing Interest}
The authors declare that they have no known competing financial interests or personal relationships that could have appeared to influence the work reported in this paper.

\section*{Acknowledgements}
This work was supported in part by the AFOSR Grant FA9550-24-1-0327.

\section*{Data Availability}
The datasets and code used in this work will be made publicly available upon acceptance. Data supporting the findings of this study are available from the corresponding author upon reasonable request.

\bibliographystyle{elsarticle-num}
\bibliography{references}

\end{document}